\documentclass[12pt,a4paper]{article}
\usepackage[latin1]{inputenc}
\usepackage{amsmath}
\usepackage{amsfonts}
\usepackage{amssymb}
\usepackage{graphicx}
\usepackage{natbib}
\usepackage{apalike}
\usepackage{multirow}
\usepackage{array}
\usepackage[ruled,linesnumbered,vlined]{algorithm2e}
\usepackage[left=2cm,right=2cm,top=2cm,bottom=2cm]{geometry}

\newcommand{\uni}{\cup} 
\newcommand{\inter}{\cap} 

\DeclareMathOperator*{\argmax}{arg\,max}

\newtheorem{definition}{Definition}

\author{Gavin Rens and Deshendran Moodley\\[12pt]
Centre for Artificial Intelligence Research,\\
University of KwaZulu-Natal, and CSIR Meraka, South Africa.
}
\title{A Hybrid POMDP-BDI Agent Architecture with\\ Online Stochastic Planning and Plan Caching}

\begin{document}
\maketitle

\begin{abstract}
\noindent
This article presents an agent architecture for controlling an autonomous agent in stochastic environments. The architecture combines the partially observable Markov decision process (POMDP) model with the belief-desire-intention (BDI) framework. The Hybrid POMDP-BDI agent architecture takes the best features from the two approaches, that is, the online
generation of reward-maximizing courses of action from POMDP theory, and sophisticated
multiple goal management from BDI theory. We introduce the advances made since the introduction of the basic architecture, including (i) the ability to pursue multiple goals simultaneously and (ii) a plan library for storing pre-written plans and for storing recently generated plans for future reuse. A version of the architecture without the plan library is implemented and is evaluated using simulations. The results of the simulation experiments indicate that the approach is feasible.
\end{abstract}

\section{Introduction}

%
%


Imagine a scenario where a planetary rover has five tasks of varying importance. The tasks could be, for instance, collecting gas (for industrial use) from a natural vent at the base of a hill, taking a temperature measurement at the top of the hill, performing self-diagnostics and repairs, reloading its batteries at the solar charging station and collect soil samples wherever the rover is. The rover is programmed to know the relative importance of collecting soil samples. The rover also has a model of the probabilities with which its various actuators fail and the probabilistic noise-profile of its various sensors. The rover must be able to reason (plan) in real-time to pursue the right task at the right time while considering its resources and dealing with various events, all while considering the uncertainties about its actions (actuators) and perceptions (sensors).

We propose an architecture for the proper control of an agent in a complex environment such as the scenario described above. The architecture combines belief-desire-intention (BDI) theory \citep{b87,rg95} and partially observable Markov decision processes (POMDPs) \citep{m82,l91}. Traditional BDI architectures (BDIAs) cannot deal with probabilistic uncertainties and they do not generate plans in real-time. A traditional POMDP cannot manage goals (major and minor tasks) as well as BDIAs can. Next, we analyse the POMDPs and BDIAs in a little more detail.

One of the benefits of agents based on BDI theory, is that they need not generate plans from scratch; their plans are already (partially) compiled, and they can act quickly once a goal is focused on. Furthermore, the BDI framework can deal with multiple goals. However, their plans are usually not optimal, and it may be difficult to find a plan which is applicable to the current situation. That is, the agent may not have a plan in its library which exactly `matches' what it ideally wants to achieve. On the other hand, POMDPs can generate optimal policies on the spot to be highly applicable to the current situation. Moreover, policies account for stochastic actions in partially observable environments. Unfortunately, generating optimal POMDP policies is usually intractable.
One solution to the intractability of POMDP policy generation is to employ a {\em continuous planning} strategy, or {\em agent-centred search} \citep{k01}. 
Aligned with agent-centred search is the \textit{forward-search} approach or \textit{online} planning approach in POMDPs \citep{rppc08}.

The traditional BDIA maintains goals as \textit{desires}; there is no reward for performing some action in some state. The reward function provided by POMDP theory is useful for modeling certain kinds of behavior or preferences. For instance, an agent based on a POMDP may want to avoid moist areas to prevent its parts becoming rusty. Moreover, a POMDP agent can generate plans which can optimally avoid moist areas. But one would not say that avoiding moist areas is the agent's \textit{task}.
And POMDP theory maintains a single reward function; there is no possibility of weighing alternative reward functions and pursuing one at a time for a fixed period---all objectives must be considered simultaneously, in one reward function. Reasoning about objectives in POMDP theory is not as sophisticated as in BDI theory.
A BDI agent cannot, however, simultaneously avoid moist areas \textit{and} collect gas; it has to switch between the two or combine the desire to avoid moist areas with every other goal. 

The Hybrid POMDP-BDI agent architecture (or HPB architecture, for short) has recently been introduced \citep{rm15a}. It combines the advantages of POMDP theoretic reasoning and the potentially sophisticated means-ends reasoning of BDI theory in a coherent agent architecture. 
In this paper, we generalize the management of goals by allowing for each goal to be pursued with different intensities, yet concurrently.

Typically, BDI agents do not deal with stochastic uncertainty. Integrating POMDP notions into a BDIA addresses this. For instance, an HPB agent will maintain a (subjective) belief state representing its probabilistic (uncertain) belief about its current state. Planning with models of stochastic actions and perceptions is possible in the HPB architecture.
The tight integration of POMDPs and BDIAs is novel to this architecture, especially in combination with desires with changing intensity levels.

This article serves to introduce two significant extensions to the first iteration \citep{rm15a} of the HPB architecture. The first extension allows for multiple intentions to be pursued simultaneously, instead of one at a time.
In the previous architecture, only one intention was actively pursued at any moment. In the new version, one agent action can take an agent closer to more than one goal at the moment the action is performed -- the result of a new approach to planning.
As a consequence of allowing multiple intentions, the policy generation module (\S~\ref{sec:Planning-by-Policy-Generation}), the desire function and the method of focusing on intentions (\S~\ref{sec:A-New-Approach-to-Focusing}) had to be adapted.
The second extension is the addition of a \textit{plan library}. Previously, a policy (conditional plan) would have to be generated periodically and regularly to supply the agent with the recommendations of actions it needs to take.
Although one of the strengths of traditional BDI theory is the availability of a plan library with pre-written plans for quick use, a plan library was excluded from the HPB architecture so as to simplify the architecture's introduction. Now we propose a framework where an agent designer can store hand-written policies in a library of plans and where generated policies are stored for later reuse. Every policy in the library is stored together with a `context' in which it will be applicable and the set of intentions which it is meant to satisfy. There are two advantages of introducing a plan library: (i) policies can be tailored by experts to achieve specific goals in particular contexts, giving the agent \textit{immediate} access to recommended courses of action in those situations, and (ii) providing a means for policies, once generated, to be stored for later reuse so that the agent can take advantage of past `experience' -- saving time and computation.

In Section~\ref{sec:Preliminaries}, we review the necessary theory, including POMDP and BDI theory.
In Section~\ref{sec:The-Basic-HPB-Architecture}, we describe the basic HPB architecture. The extensions to the basic architecture are presented in Section~\ref{The-Extended-HPB-Architecture}.
Section~\ref{sec:Simulations} describes two simulation experiments in which the proposed architecture is tested, evaluating the performance on various dimensions.
The results of the experiments confirm that the approach may be useful in some domains.
The last section discusses some related work and points out some future directions for research in this area.

\section{Preliminaries}
\label{sec:Preliminaries}

The basic components of a BDI architecture \citep{w99,w02} are
\begin{itemize}
\itemsep=0pt
\item a set or knowledge-base $B$ of beliefs;
\item an option generation function $\mathit{wish}$, generating the objectives the agent would ideally like to pursue (its desires);
\item a set of desires $D$ (goals to be achieved);
\item a `focus' function which selects intentions from the set of desires;
\item a structure of intentions $I$ of the most desirable options/desires returned by the focus function;
\item a library of plans and subplans;
\item a `reconsideration' function which decides whether to call the focus function;
\item an execution procedure, which affects the world according to the plan associated with the intention;
\item a sensing or perception procedure, which gathers information about the state of the environment; and
\item a belief update function, which updates the agent's beliefs according to its latest observations and actions.
\end{itemize}
Exactly how these components are implemented result in a particular BDI architecture.

\begin{algorithm}[t]
\caption{Basic BDI agent control loop}
\label{algo:1}
\KwIn{$B_0$: initial beliefs} 
\KwIn{$I_0$: initial intentions}
$B \gets B_0$\;
$I \gets I_0$\;
$\pi \gets null$ \;
\While{alive}{
$p \gets \mathit{getPercept}()$\;
$B \gets \mathit{update}(B,p)$\;
$D \gets \mathit{wish}(B,I)$\;
$I \gets \mathit{focus}(B,D,I)$\;
$\pi \gets \mathit{plan}(B,I)$\;
$\mathit{execute}(\pi)$\;
}
\end{algorithm}

Algorithm~\ref{algo:1} (adapted from \citet[Fig.~2.3]{w00}) is a basic BDI agent control loop.
$\pi$ is the current plan to be executed.
$\mathit{getPercept}(\cdot)$ senses the environment and returns a percept (processed sensor data) which is an input to $\mathit{update}(\cdot)$, which updates the agent's beliefs.
$\mathit{wish}: B \times I \to D$ generates a set of desires, given the agent's beliefs, current intentions and possibly its innate motives. It is usually impractical for an agent to pursue the achievement of all its desires. It must thus filter out the most valuable and achievable desires. This is the function of $\mathit{focus}: B \times D \times I \to I$, taking beliefs, desires and current intentions as parameters. Together, the processes performed by $\mathit{wish}$ and $\mathit{focus}$ may be called deliberation, formally encapsulated by the $\mathit{deliberate}$ procedure.
 $\mathit{plan}(\cdot)$ returns a plan from the plan library to achieve the agent's current intentions.

A more sophisticated controller would have the agent consider {\em whether} to re-deliberate, with a $\mathit{reconsider}$ function placed just before deliberation would take place.
%
The agent could also test at every iteration through the main loop whether the currently pursued intention is still possibly achievable.
Serendipity could also be taken advantage of by periodically testing whether the intention has been achieved, without the plan being fully executed. Such an agent is considered `reactive' because it executes one action per loop iteration; this allows for deliberation between executions.
%
There are various mechanisms which an agent might use to decide when to reconsider its intentions.
See, for instance, \citet{b87,pr90,kg91,kg92,sw00,sw01b,swp04}.

\bigskip
In a partially observable Markov decision process (POMDP), the actions the agent performs have non-deterministic effects in the sense that the agent can only predict with a likelihood in which state it will end up after performing an action. Furthermore, its perception is noisy. That is, when the agent uses its sensors to determine in which state it is, it will have a probability
distribution over a set of possible states to reflect its conviction for being in each state.

Formally \citep{klc98}, a POMDP is a tuple
$\langle S,A,T,R,Z,P,b^0 \rangle$ with
\begin{itemize}
\itemsep=0pt
\item $S$, a finite set of states of the world (that the agent can be in), 
\item $A$ a finite set of actions (that the agent can choose to execute), 
\item a transition function $T(s,a,s')$, the probability of being in $s'$ after performing action $a$ in state $s$, 
\item $R(a,s)$, the immediate reward gained for executing action $a$ while in state $s$, 
\item $Z$, a finite set of observations the agent can perceive in its world, 
\item a perception function $P(s',a,z)$, the probability of observing $z$ in state $s'$ resulting from performing action $a$ in some other state, and 
\item $b^0$ the initial probability distribution over all states in
$S$.
\end{itemize} 

In general, we regard an observation as the signal recognized by a sensor; the signal is generated by some event which is not directly perceivable.

A belief state $b$ is a set of pairs $\langle s,p\rangle$ where each state $s$ in $b$ is associated with a probability $p$. All probabilities must sum up to one, hence, $b$ forms a probability distribution over the set $S$ of all states. To update the agent's beliefs about the world, a special function $\mathit{SE}(z, a, b)=b_n$ is defined as
\begin{equation}
b_n(s') = \frac{P(s',a,z)\sum_{s \in S}T(s,a,s')b(s)}{Pr(z | a,b)},
\label{eq:belupdate}
\end{equation}
where $a$ is an action performed in `current' belief state $b$, $z$ is the resultant observation and $b_n(s')$ denotes the probability of the agent being in state $s'$ in `new' belief state $b_n$. Note that $Pr(z\,|\,a,b)$ is a normalizing constant.

Let the {\em planning horizon} $h$ (also called the {\em look-ahead
  depth}) be the number of future steps the agent plans ahead each
time it plans. $V^*(b,h)$ is the {\em optimal} value
of future courses of actions the agent can take with respect to a finite
horizon $h$ starting in belief state $b$. This function assumes that
at each step the action that will maximize the state's value will be
selected.

Because the reward function $R(a,s)$ provides feedback about the utility of a particular state $s$ (due to $a$ executed in it), an agent who does not know in which state it is in cannot use this reward function directly. The agent must consider, for each state $s$, the probability $b(s)$ of being in $s$, according to its current belief state $b$. Hence, a \textit{belief} reward function $\rho(a,b)$ is defined, which takes a belief state as argument.
Let $\rho(a,b):=\sum_{s\in S}R(a,s)b(s)$.

The optimal \textit{state-value} function is define by
\[
V^*(b,h) := \max_{a\in\mathcal{A}}\Big[\rho(a,b) + \gamma\sum_{z\in Z}Pr(z\,|\, a,b)  V^*(\mathit{SE}(z,a,b),h-1)\Big],
\]
where $0 \leq\gamma < 1$ is a factor to discount the value of future
rewards and $Pr(z\,|\, a,b)$ denotes the probability of reaching
belief state $b_n = \mathit{SE}(z,a,b)$.  
While $V^*$ denotes the optimal value of a belief state, function $Q^*$ denotes the optimal \textit{action-value}:
\[
Q^*(a,b,h) := \rho(a,b) + \gamma\sum_{z\in Z}Pr(z\,|\, a,b)V^*(\mathit{SE}(z,a,b),h-1)
\]
is the value of executing $a$ in the current belief state, plus the total expected value of belief states reached thereafter.

\section{The Basic HPB Architecture}
\label{sec:The-Basic-HPB-Architecture}

In BDI theory, one of the big challenges is to know \textit{when} the agent should switch its current goal and \textit{what} its new goal should be \citep{swp04}. To address this challenge, we propose that an agent should maintain intensity levels of desire for every goal.
This intensity of desire could be interpreted as a kind of emotion.
The goals most intensely desired should be the goals sought (the agent's intentions). We also define the notion of how much an intention is satisfied in the agent's current belief state.
For instance, suppose that out of five possible goals, the agent currently most desires to watch a film and to eat a snack. Then these two goals become the agent's \textit{intentions}. However, eating is not allowed inside the film-theatre, and if the agent were to go buy a snack it would miss the beginning of the film. So the total reward for first watching the film then buying and eating a snack is higher than first eating then watching.
As soon as the film-watching goal is satisfied, it is no longer an intention. But while the agent was watching the film, the desire-level of the (non-intention) goal of being at home has been increasing. However, it cannot become an intention because snack-eating has not yet been satisfied. Going home cannot simply become an intention and dominate snack-eating, because the architecture is designed so that current intentions have precedence over non-intention goals, else there is a danger that the agent will vacillate between which goals to pursue. Nonetheless, snack-eating may be ejected from the set of intentions under the special condition that the agent is having an unusually hard time achieving it. For instance, if someone stole its wallet in the theatre, the agent can no longer have the current intention (i.e., actively pursue) eating a snack. Hence, in our architecture, if an intention takes `too long' to satisfy, it is removed from the set of intentions. As soon as the agent gets home or is close to home, the snack-eating goal will probably become an intention again and the agent will start making plans to satisfy eating a snack. Moreover, the desire-level of snack-eating will now be very high (it has been steadily increasing) and the agent's actions will be biased towards satisfying this intention over other current intentions (e.g., over getting home, if it is not yet there).

\begin{figure*}
\centering
\includegraphics[scale=0.45]{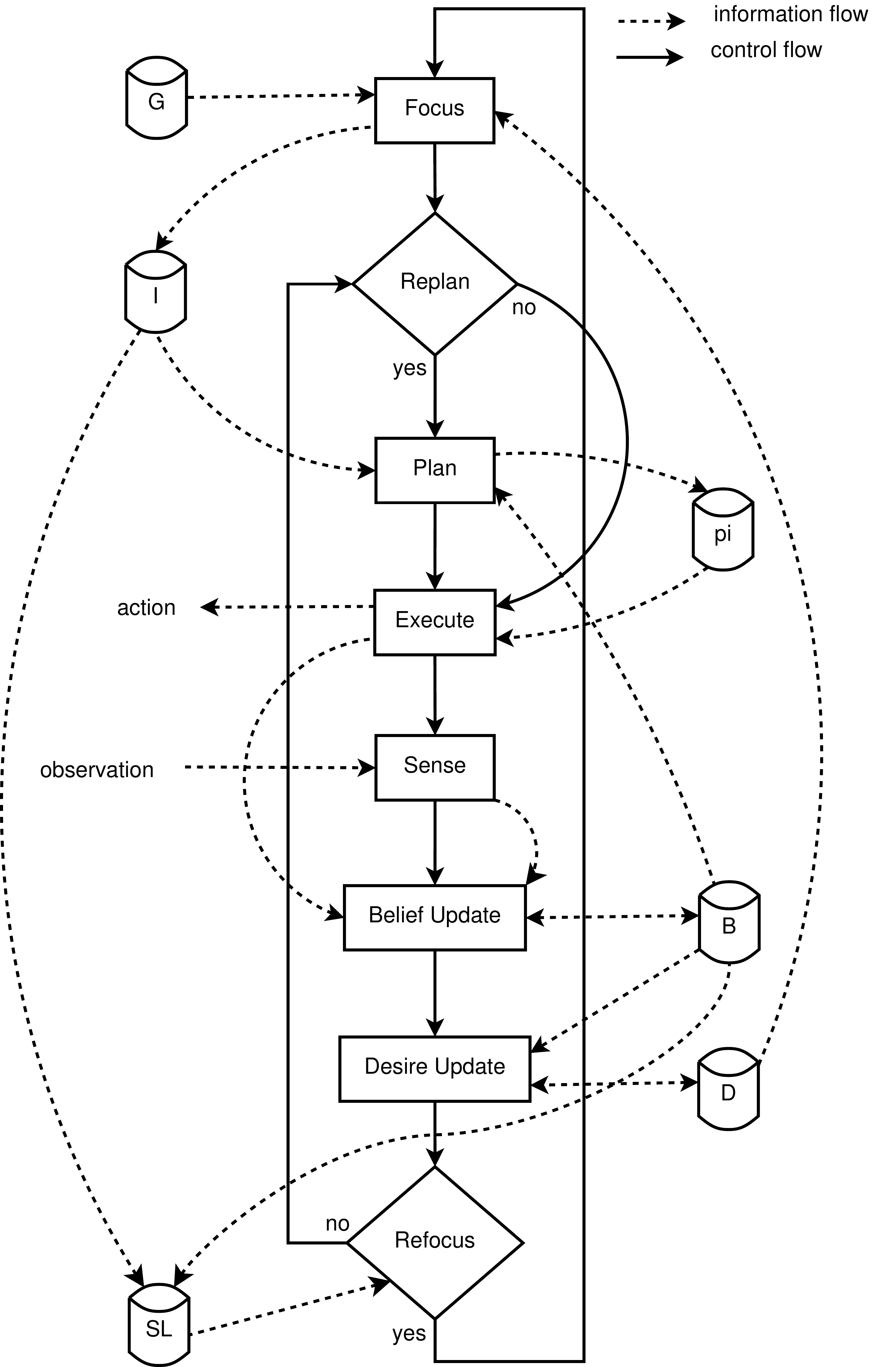}
\vspace{5mm}
\caption{Operational semantics of the basic HPB architecture. SL stands for $\mathit{Satf\_levels}$. Note that $\mathit{Satf\_levels}$ depends on the current belief state and not on desire levels. Planning is also independent of desire levels. The focus function depends on desire levels and on satisfaction levels. In the case of plans consisting of a single action, the Replan decision node always returns `yes'.}
\label{fig:hpbagentdiagram}
\end{figure*}

A Hybrid POMDP-BDI (HPB) agent \citep{rm15a} maintains (i) a belief state which is periodically updated, (ii) a mapping from goals to numbers representing the level of desire to achieve the goals, and (iii) the current set of intentions, the goals with the highest desire levels (roughly speaking). As the agent acts, its desire levels are updated and it may consider choosing new intentions and discard others based on new desire levels. Refer to Figure~\ref{fig:hpbagentdiagram} for an overview of the operational semantics. The figure refers to concepts defined in the following subsection.

\subsection{Declarative Semantics}

The \textit{state} of an HPB agent is defined by the tuple $\langle B, D, I \rangle$, where $B$ is the agent's current belief state (i.e., a probability distribution over the states $S$, defined below), $D$ is the agent's current desire function and $I$ is the agent's current intention. More will be said about $D$ and $I$ a little later.

An HPB agent could be defined by the tuple $\langle \mathit{Atrb}, G, A, Z, T, P, \mathit{Util}\rangle$, where

\begin{itemize}
\item
$\mathit{Atrb}$ is a set of attribute-sort pairs (for short, the \textit{attribute set}). For every $(\mathit{atrb}:\mathit{sort})\in \mathit{Atrb}$, $\mathit{atrb}$ is the name or identifier of an attribute of interest in the domain of interest, like $\mathtt{BatryLevel}$ or $\mathtt{Direction}$, and $\mathit{sort}$ is the set from which $\mathit{atrb}$ can take a value, for instance, real numbers in the range $[0,55]$ or a list of values like $\{\mathtt{North}$, $\mathtt{East}$, $\mathtt{West}$, $\mathtt{South}\}$. So $\{(\mathtt{BatryLevel}:[0,55]),(\mathtt{Direction}:$ $\{\mathtt{North},\mathtt{East},\mathtt{West},\mathtt{South}\})\}$ could be an attribute set.

A state $s$ is induced from $\mathit{Atrb}$ as one possible way of assigning values to attributes: $s=\{(\mathit{atrb}:v)\mid (\mathit{atrb}:\mathit{sort})\in \mathit{Atrb}, v\in \mathit{sort},$ if $(\mathit{atrb}:v),(\mathit{atrb}':v')\in s$ and $\mathit{atrb}=\mathit{atrb}'$, then $v=v'\}$.
The set of all possible states is denoted $S$.

\item
$G$ is a set of goals. 
A goal is a subset of some state $s\in S$.
For instance, 
$\{(\mathtt{BatryLevel} : 13),(\mathtt{Direction} : \mathtt{South})\}$ is a goal, and so are
$\{(\mathtt{BatryLevel} :33)\}$ and $\{(\mathtt{Direction} : \mathtt{West})\}$.
The set of goals is given by the agent designer as `instructions' about the agent's tasks.

\item
$A$ is a finite set of actions.

\item
$Z$ is a finite set of observations.

\item
$T$ is the transition function of POMDPs.

\item
$P$ is the perception function of POMDPs.

\item
$\mathit{Util}$ consists of two functions $\mathit{Pref}$ and $\mathit{Satf}$ which allow an agent to determine the utilities of alternative sequences of actions. $\mathit{Util}=\langle \mathit{Pref},\mathit{Satf} \rangle$.

$\mathit{Pref}$ is the preference function with a range in $\mathbb{R}\inter[0,1]$. It takes an action $a$ and a state $s$, and returns the preference (any real number) for performing $a$ in $s$. That is, $\mathit{Pref}(a,s)\in[0,1]$. Numbers closer to 1 imply greater preference and numbers closer to 0 imply less preference. Except for the range restriction of $[0,1]$, it has the same definition as a POMDP reward function, but its name indicates that it models the agent's preferences and not what is typically thought of as rewards. An HPB agent gets `rewarded' by achieving its goals. The preference function is especially important to model action costs; the agent should prefer `inexpensive' actions. $\mathit{Pref}$ has a local flavor.
Designing the preference function to have a value lying in [0,1] may sometimes be challenging, but we believe it is always possible.

$\mathit{Satf}$ is the satisfaction function with a range in $\mathbb{R}\inter[0,1]$. It takes a state $s$ and an intention $I$, and returns a value representing the degree to which the state satisfies the intention. That is, $\mathit{Satf}(I,s)\in[0,1]$. It is completely up to the agent designer to decide how the satisfaction function is defined, as long as numbers closer to 1 mean more satisfaction and numbers closer to 0 mean less satisfaction. $\mathit{Satf}$ has a global flavor.
\end{itemize}

Figure~\ref{fig:hpbagentdiagram} shows a flow diagram representing the operational semantics of the basic HPB architecture.

\subsection{The Desire Function}

The desire function $D$ is a total function from goals in $G$ into the positive real numbers $\mathbb{R}^+$. The real number represents the intensity or level of desire of the goal. For instance, $(\{(\mathit{BatryLevel}:13),(\mathit{WeekDay}:\mathit{Tue})\},2.2)$ could be in $D$, meaning that the goal of having the battery level at 13 and the week-day Tuesday is desired with a level of 2.2. $(\{(\mathit{BatryLevel}:33)\},56)$ and $(\{(\mathit{WeekDay}:\mathit{Wed})\},444)$ are also examples of desires in $D$.

$I$ is the agent's current intention; an element of $G$; the goal with the highest desire level. This goal will be actively pursued by the agent, shifting the importance of the other goals to the background.
The fact that only one intention is maintained makes the HPB agent architecture quite different to standard BDIAs.

We propose the following desire update rule.
\begin{equation}
D(g) \gets D(g) + 1 - \mathit{Satf}_\beta(g,B)
\label{eq:DU}
\end{equation}
Rule~\ref{eq:DU} is defined so that as $\mathit{Satf}_\beta(g,B)$ tends to one (total satisfaction), the intensity with which the incumbent goal is desired does not increase. On the other hand, as $\mathit{Satf}_\beta(g,B)$ becomes smaller (more dissatisfaction), the goal's intensity is incremented. 
The rule transforms $D$ with respect to $B$ and $g$. A goal's intensity should drop the more it is being satisfied. 
The update rule thus defines how a goal's intensity changes over time with respect to satisfaction.

Note that desire levels never decrease. This does not reflect reality. It is however convenient to represent the intensity of desires like this: only \textit{relative} differences in desire levels matter in our approach and we want to avoid unnecessarily complicating the architecture.

\subsection{Focusing and Satisfaction Levels}

$\mathit{Focus}$ is a function which returns one member of $G$ called the (current) intention $I$.
In the initial version of the architecture, the goal selected is the one with the highest desire level.
After every execution of an action in the real-world, $\mathit{Refocus}$ is called to decide whether to call $\mathit{Focus}$ to select a new intention.
$\mathit{Refocus}$ is a meta-reasoning function analogous to the $\mathit{reconsider}$ function mentioned in Section~\ref{sec:Preliminaries}.
It is important to keep the agent focused on one goal long enough to give it a reasonable chance of achieving it. It is the job of $\mathit{Refocus}$ to recognize when the current intention seems impossible or too expensive to achieve.

Let $\mathit{Satf\_levels}$ be the sequence of satisfaction levels of the current intention since it became active and let $\mathit{MRY}$ be a designer-specified number representing the length of a sub-sequence of $\mathit{Satf\_levels}$---the $\mathit{MRY}$ last satisfaction levels.

One possible definition of $\mathit{Refocus}$ is
\begin{equation*}
\mathit{Refocus}(c,\theta) \stackrel{\mathit{def}} = \left\{
\begin{array}{rl}
\mbox{`no'} & \text{if } |\mathit{Satf\_levels}| < \mathit{MRY}\\
\mbox{`yes'} & \text{if } c < \theta\\
\mbox{`no'} & \text{otherwise,}
\end{array} \right.
\end{equation*}
where $c$ is the average change from one satisfaction level to the next in the agent's `memory' $\mathit{MRY}$, and $\theta$ is some threshold, for instance, $0.05$. If the agent is expected to increase its satisfaction by at least, say, 0.1 on average for the current intention, then $\theta$ should be set to 0.1. With this approach, if the agent `gets stuck' trying to achieve its current intention, it will not blindly keep on trying to achieve it, but will start pursuing another goal (with the highest desire level).
Some experimentation will likely be necessary for the agent designer to determine a good value for $\theta$ in the application domain.

Note that if an intention was not well satisfied, its desire level still increases at a relatively high rate. So whenever the agent focuses again, a goal not well satisfied in the past will be a top contender to become the intention (again).

\subsection{Planning for the Next Action}

A basic HPB agent controls its behaviour according to the policies it generates. $\mathit{Plan}$ is a procedure which generates a POMDP policy $\pi$ of depth $h$.
Essentially, we want to consider all action sequences of length $h$ and the belief states in which the agent would find itself if it followed the sequences. Then we want to choose the sequence (or at least its first action) which yields the least cost and which ends in the belief state most satisfying with respect to the intention.

Planning occurs over an agents belief states. The satisfaction and preference functions thus need to be defined for belief states:
The satisfaction an agent gets for an intention in its current belief state is defined as
\[
\mathit{Satf}_\beta(I,B) := \sum_{s\in S}\mathit{Satf}(I,s)B(s),
\]
where $\mathit{Satf}(I,s)$ is defined above and $B(s)$ is the probability of being in state $s$.
The definition of $\mathit{Pref}_\beta$ has the same form as the reward function $\rho$ over belief states in POMDP theory:
\[
\mathit{Pref}_\beta(a,B) := \sum_{s \in S}\mathit{Pref}(a,s)B(s),
\]
where $\mathit{Pref}(a,s)$ was discussed above.

During planning, preferences and intention satisfaction must be maximized.
The main function used in the $\mathit{Plan}$ procedure is the HPB action-state value function $Q^*_\mathit{HPB}$, giving the value of some action $a$, conditioned on the current belief state $B$, intention $I$ and look-ahead depth $h$:
\begin{align*}
& Q^*_\mathit{HPB}(a,B,I,h):=\alpha\mathit{Satf}_\beta(I,B) + (1-\alpha)\mathit{Pref}_\beta(a,B)\\
&\qquad + \gamma\sum_{z\in Z}Pr(z\mid a,B) \max_{a'\in A}Q^*_\mathit{HPB}(a',B',I,h-1),\\
& Q^*_\mathit{HPB}(a,B,I,1) :=\alpha\mathit{Satf}_\beta(I,B) + (1-\alpha)\mathit{Pref}_\beta(a,B),
\end{align*}
where $B'=\mathit{SE}(a,z,B)$, $0\leq\alpha\leq1$ is the goal/preference `trade-off' factor, $\gamma$ is the normal POMDP discount factor and $\mathit{SE}$ is the normal POMDP state estimation function.

$\mathit{Plan}$ returns $\argmax_{a\in A}Q^*_\mathit{HPB}(a,B,I,h)$, the trivial policy of a single action.

\section{The Extended HPB Architecture}
\label{The-Extended-HPB-Architecture}

The operational semantics of the extended architecture is essentially the same as for the first version, except that a plan library is now involved. The agent starts off with an initial set of intentions, a subset of its goals. For the current set of intentions, it must either select a plan from the plan library or generate a plan to pursue all its intentions. At every iteration of the agent's control loop, an action is performed, an observation is made, the belief state is updated, and a decision is made whether to modify the set of intentions. But only when the current policy (conditional plan) is `exhausted' does the agent seek a new policy, by consulting its plan library, and if an adequate policy is not found, generating one.

In the next subsection, we introduce some new notation and changes made to the architecture.
Section~\ref{sec:A-New-Approach-to-Focusing} discusses how the focussing procedure must change to accommodate the changes.
Section~\ref{sec:Planning-by-Policy-Generation} explains how policies are generated for simultaneous pursuit of multiple goals.
Finally, Section~\ref{sec:Introducing-a-Plan-Library} presents the plan library, which was previously unavailable, and how the agent and agent designer can use it to their benefit.

\subsection{Prologue}
The HPB agent model gets three new component -- a goal weight function $W$, a compatibility function $\mathit{Cpbl}$ and the plan library $\mathit{Lbry}$. It can thus be defined by the tuple $\langle \mathit{Atrb}$, $G$, $W$, $\mathit{Cpbl}$, $A$, $Z$, $T$, $P$, $\mathit{Util}$, $\mathit{Lbry}\rangle$.

In the previous version, satisfaction and preference were traded-off by ``trade-off factor'' which was not explicitly mentioned in the agent model. Actually the trade-off factor should have been part of the model, because it must be provided by the agent designer, and it directly affects the agent's behaviour.
In the new version, every goal $g\in G$ will be weighted by $W(g)$ according to the importance of $g$ to the agent.
Goal weights are constrained such that
$W(g)>0$ for all $g\in G$, and $\sum_{g\in G}W(g)=1$.

The third fundamental extension is that $I$ becomes a \textit{set} of intentions. In this way, an HPB agent may actively pursue several goals simultaneously.
For example, a planetary rover may want to travel to its recharging station and simultaneously make same atmospheric measurements en route.

The first version has also been changed so that the set of goals $G$ is simply a set of names, rather than restricting a goal to be a set of attribute values, as was previously done. Goals are defined by how they are used in the architecture, particularly by their involvement in the definition of satisfaction functions.

In the extended architecture, it will be convenient to use more compact notation: Here we let $\mathit{Util}=\langle\kappa,\sigma\rangle$, where $\kappa$ is the same as $\mathit{Pref}$ and $\sigma$ is a \textit{set} of satisfaction functions $\{\sigma^g\mid g\in G, \sigma^g=\mathit{Satf}(g)\}$. In particular, we move away from a \textit{preference} function, and rather think of a \textit{cost} function $\kappa$. Preferences will be captured by the set of satisfaction functions.

As a consequence of being able to pursue several goals at the same time, there exists a danger that the agent will pursue one intention when it necessarily causes another intention to become less satisfied.
For instance, visiting the USA regional headquarters is diametrically opposite to visiting the China regional headquarters at the same time.
Other examples of goals which should be `disjoint' are $\mathtt{work{-}in{-}garden}$ and $\mathtt{have{-}lunch}$, and $\mathtt{recharge{-}battery}$ and $\mathtt{replace{-}battery}$.
The solution we use is to list, for each goal $g\in G$, all other goals which are \textit{compatible} with it, in the sense that their simultaneous pursuit `effective' (defined by the agent designer).
Let $\mathit{Cpbl}(g)$ denote the set of goals compatible with $g$. It is mandatory that 
$g\in\mathit{Cpbl}(g)$. Two goals $g$ and $g'$ are called \textit{incompatible} if and only if $g'\not\in\mathit{Cpbl}(g)$ or $g\not\in\mathit{Cpbl}(g')$.

Suppose $G=\{\mathtt{visit{-}USA{-}HQ}$, $\mathtt{visit{-}China{-}HQ}$, $\mathtt{work{-}in{-}garden}$, $\mathtt{have{-}lunch}$,\\ $\mathtt{recharge{-}battery}$, $\mathtt{replace{-}battery}\}$.
Then an agent designer may specify
\[\mathit{Cpbl}(\mathtt{visit{-}USA{-}HQ}) = \{\mathtt{visit{-}USA{-}HQ},\mathtt{have{-}lunch}\}\] and \[\mathit{Cpbl}(\mathtt{recharge{-}battery}) = \{\mathtt{recharge{-}battery},\mathtt{work{-}in{-}garden}, \mathtt{have{-}lunch}\}.\]
Note that $\mathtt{work{-}in{-}garden}$ and $\mathtt{have{-}lunch}$ are incompatible.

\subsection{A New Approach to Focusing}
\label{sec:A-New-Approach-to-Focusing}

Given that $I$ is a \textit{set} of intentions, ensuring that the `correct' goals are intentions at the `right' time to ensure that the agent behaves as desired, requires some careful thought.
It is still important to keep the agent focused on one intention long enough to give it a reasonable chance of achieving it, temporarily stop pursuing intentions it is struggling to achieve.

The HPB architecture does not have a focus \textit{function} which returns a subset of $G$ of intentions $I$.
Rather, we have a set of procedures which decide at each iteration which intention to remove from $I$ (if any) and which goal to add to $I$ (if any). Incompatible must also be dealt with.

Let $\mathit{Satf\_levels}(g)$ be the sequence of satisfaction levels of some goal $g\in I$ since $g$ became active (i.e., was added to $I$) and let $\mathit{MRY}$ be a
number representing the length of a sub-sequence of $\mathit{Satf\_levels}(g)$---the $\mathit{MRY}$ last satisfaction levels of goal $g$.
$\mathit{Remove}$ is defined exactly like $\mathit{Refocus}$:
\begin{equation*}
\mathit{Remove}(g,I) := \left\{
\begin{array}{rl}
\mbox{`no'} & \text{if } |\mathit{Satf\_levels(g)}| < \mathit{MRY}(g)\\
\mbox{`yes'} & \text{if } \delta(g) < \theta\\
\mbox{`no'} & \text{otherwise,}
\end{array} \right.
\end{equation*}
where $\delta(g)$ is the average change from one satisfaction level of $g$ to the next in the agent's `memory', and $\theta$ is the threshold above which $\delta(g)$ must be for $g$ to remain an intention.

Let $\mathit{MI}$ be the currently \textit{most intense} goal defined as
\[
\mathit{MI} := \argmax_{g\in G}D(g).
\]

We define two focusing strategies for \textit{sets} of intentions: the over-optimistic strategy and the compatibility strategy.

\subsubsection{Over-optimistic Strategy}

This strategy ignores compatibility issues between goals. In this sense, the agent is (over) optimistic that it can successfully simultaneously pursue goals which are incompatible.

Add $\mathit{MI}$ to $I$ only if $\mathit{MI}\not\in I$.
If $\mathit{MI}$ is added to $I$, clear $\mathit{MI}$'s record of satisfaction levels, that is, let $\mathit{Satf\_levels}(\mathit{MI})$ be the empty sequence.

Next:
For every $g\in I$, if $|I|>1$ and $\mathit{Remove}(g,I)$ returns `yes', then remove $g$ from $I$.

\subsubsection{Compatibility Strategy}


Add $\mathit{MI}$ to $I$ only if $\mathit{MI}\not\in I$ and there does not exists a $g\in I$ such that $g\not\in \mathit{Cpbl}(\mathit{MI})$.
If $\mathit{MI}$ is added to $I$, clear $\mathit{MI}$'s record of satisfaction levels, that is, let $\mathit{Satf\_levels}(\mathit{MI})$ be the empty sequence.

Next:
For every $g\in I$, if $|I|>1$ and $\mathit{Remove}(g,I)$ returns `yes', then remove $g$ from $I$.

There is one case which must still be dealt with in the compatibility strategy: Suppose for some $g\in G$, $\bar{g}\not\in\mathit{Cpbl}(g)$. Further suppose that $I=\{\bar{g}\}$ (i.e., $|I|=1$) and $g$ is and remains the most intensely desired goal. Now, $g$ may not be added to $I$ because it is incompatible with $\bar{g}$, no other goal will be attempted to be added to $I$ and $\bar{g}$ may not be removed while it is the only intention, \textit{even if} $\mathit{Remove}(\bar{g},I)$ returns `yes'. What could easily happen in this case is that $g$ will continually increase in desire level, $\bar{g}$'s average satisfaction level will remain below the change threshold (i.e., $\delta(\bar{g}) < \theta$ remains true), and the agent continues to pursue only $\bar{g}$.
To remedy this `locked' situation, the following procedure is run after the previous `add' and `remove' procedures are attempted.
If $I=\{\bar{g}\}$, $\bar{g}\not\in \mathit{Cpbl}(\mathit{MI})$ and $\mathit{Remove}(\bar{g},I)$ returns `yes', then remove $\bar{g}$ from $I$, add $\mathit{\mathit{MI}}$ to $I$ and clear $\mathit{MI}$'s record of satisfaction levels.

\subsubsection{A New Desire Function}

The old rule (in new notation) is still available:
\begin{equation}
D(g) \gets D(g) + W(g)(1 - \sigma^g_\beta(B)).
\label{eq:DU2}
\end{equation}

We have found through experimentation that when an intention-goal's desire levels are updated, non-intention-goals may not get the opportunity to become intentions. In other words, it may happen that whenever new non-intention-goals are considered to become intentions, they are always `dominated' by goals with higher levels of desire which are already intentions. By disallowing intentions' desire levels to increase, non-intentions get the opportunity to `catch up' with their desire levels.
A new form of the desire update rule is thus proposed for this version of the architecture:
\begin{equation}
D(g) \gets D(g) + (1-i(I,g))W(g)(1 - \sigma^g_\beta(B))
\label{eq:DU1}
\end{equation}
The term $(1-i(I,g))$ in \eqref{eq:DU1} ensures that a goal's desire level changes if and only if the goal is not an intention.

Both forms of the rule are defined so that as $\sigma^g_\beta(B)$ tends to one (total satisfaction), the intensity with which the incumbent goal is desired does not increase. On the other hand, as $\sigma^g_\beta(B)$ becomes smaller (more dissatisfaction), the goal's intensity is incremented---by at most its weight of importance $W(g)$. 
A goal's intensity should drop the more it is being satisfied. 

However, update rule \eqref{eq:DU2} which is independent of whether a goal is an intention may still result in better performance in particular domains. (This question needs more research.) It is thus left up to the agent designer to desire which form of the rule better suits the application domain.

\subsection{Planning by Policy Generation}
\label{sec:Planning-by-Policy-Generation}

In this section, we shall see how the planner can be extended to compute a policy which pursues several goals simultaneously. Goal weights are also incorporated into the action-state value function.

The satisfaction an agent gets for an intention $g$ at its current belief state is defined as
\[
\sigma^g_\beta(B) := \sum_{s\in S}\sigma^g(s)B(s),
\]
where $\sigma^g(s)$ is defined above and $B(s)$ is the probability of being in state $s$.
The definition of $\kappa_\beta$ has the same form as the reward function $\rho$ over belief states in POMDP theory:
\[
\kappa_\beta(a,B) := \sum_{s \in S}\kappa(a,s)B(s),
\]
where $\kappa(a,s)$ was discussed above.

The main function used in the $\mathit{Plan}$ procedure is the HPB action-state value function $Q^*_\mathit{HPB}$, giving the value of some action $a$, conditioned on the current belief state $B$
and look-ahead depth $h$:
\begin{align*}
& Q^*_\mathit{HPB}(a,B,I,h):= i(I,g_1)W(g_1)\sigma^{g_1}_\beta(B) + \cdots + i(I,g_n)W(g_n)\sigma^{g_n}_\beta(B) - \kappa_\beta(a,B)\\
&\qquad + \gamma\sum_{z\in Z}Pr(z\mid a,B) \max_{a'\in A}Q^*_\mathit{HPB}(a',B',I,h-1),\\
& Q^*_\mathit{HPB}(a,B,I,1) := i(I,g_1)W(g_1)\sigma^{g_1}_\beta(B) + \cdots + i(I,g_n)W(g_n)\sigma^{g_n}_\beta(B) - \kappa_\beta(a,B),
\end{align*}
where
\begin{itemize}
\itemsep=0pt
\item $i(I,g_j)=1$ if $g_j\in I$, else $i(I,g_j)=0$ if $g_j\not\in I$,
\item $\langle g_1,\ldots,g_n\rangle$ is an ordering of the goals in $G$,
\item $\sigma^{g}_\beta(\cdot)$ and $\kappa_\beta(\cdot)$ are the expected (w.r.t.\ a belief state) values of $\sigma^{g}(\cdot)$, resp., $\kappa(\cdot)$,
\item $B'=\mathit{SE}(a,z,B)$,
\item $\gamma$ is the normal POMDP discount factor and
\item $\mathit{SE}$ is the normal POMDP state estimation function.
\end{itemize}

Now, instead of $\mathit{Plan}$ returning a single action (assuming $h>1$), $\mathit{Plan}$ generates a tree-structures plan of depth $h$, conditioned on observations, that is, a \textit{policy}.
With a policy of depth $h$, an agent can execute a sequence of $h$ actions, where the choice of exactly which action to take at each step depends on the observation received just prior. 
$\argmax_{a\in A}$ $Q^*_\mathit{HPB}(a,B,I,h)$ is used at every choice point to construct the policy.

\begin{figure}[t]
\label{fig:policy}
\centering
\includegraphics[scale=0.6]{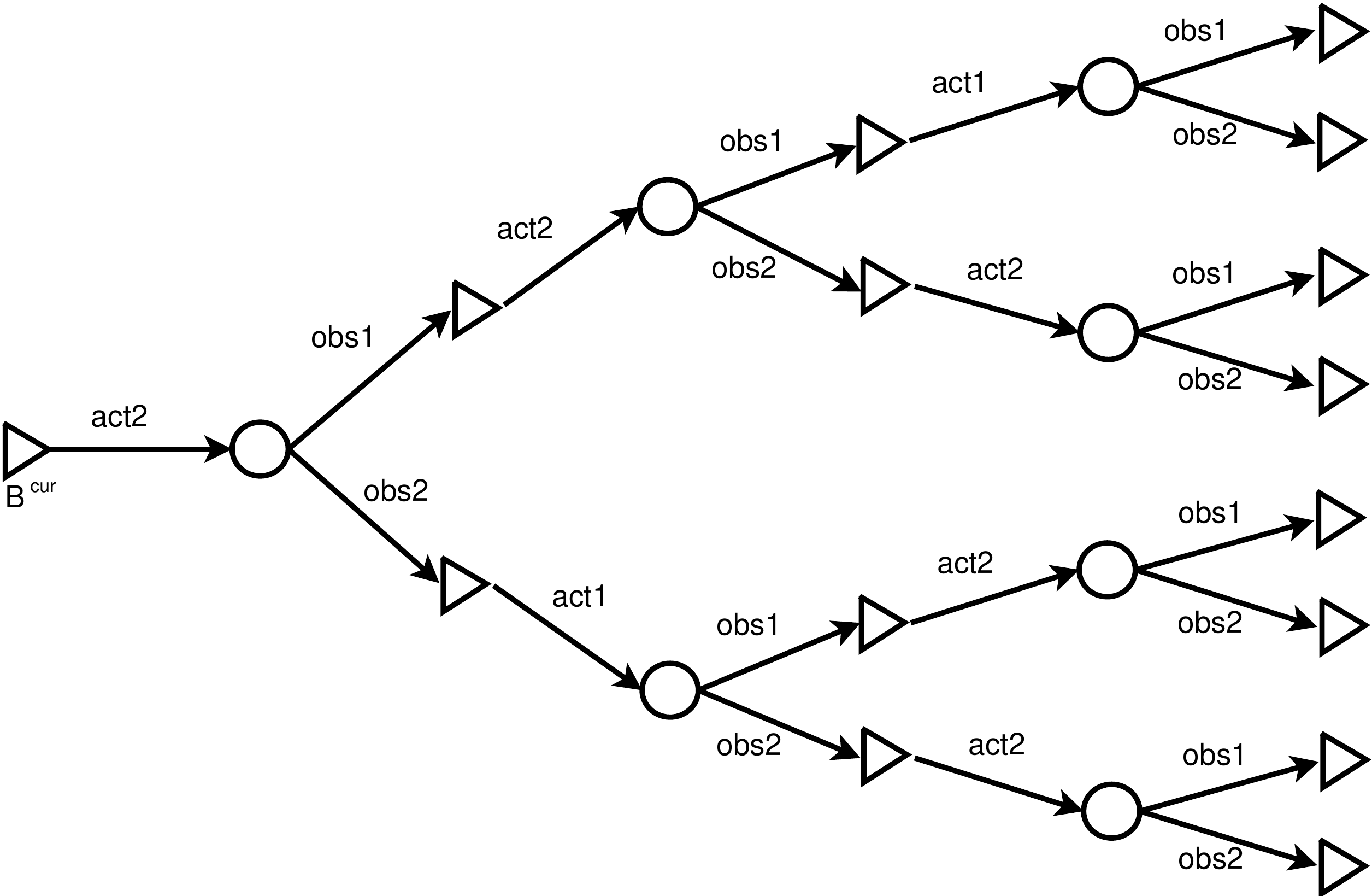}
\caption{An example policy of depth 3.}
\end{figure}

Figure~\ref{fig:policy} is a graphical example of a policy with two actions and two observations. The agent is assumed to be in belief state $B^\mathit{cur}$ when the policy is generated. At every belief state node (triangles), the optimal action is recommended. After an action is performed, all/both observations are possible and thus considered. There is thus a choice at every $\circ$ node; however, it is not a choice for the agent, rather, it is a choice for the environment which observation to send to the agent. Given the action performed, for every possible observation, a different belief state is generated. At every $\triangleright$ node (belief state), $\argmax_{a\in A}$ $Q^*_\mathit{HPB}(a,\triangleright,I,h)$ is applied to determine the action to perform there. (In theory, the agent can choose to perform any action at these $\triangleright$ nodes, but our agent will take the recommendations of POMDP theory for optimal behavior.)
The agent will perform $\mathtt{act2}$ first, then depending on whether $\mathtt{obs1}$ or $\mathtt{obs2}$ in sensed, the agent should next (according to the policy) perform $\mathtt{act2}$, respectively, $\mathtt{act1}$. Then a third action will be performed according to the policy and conditional on which observation is sensed.

\subsection{Introducing a Plan Library}
\label{sec:Introducing-a-Plan-Library}

Another extension of the basic architecture is that a language based on the attributes is introduced.
The language $L$ is the set of all \textit{sentences}. Let $\phi$ and $\psi$ be sentences. Then the following are also sentences.
\begin{itemize}
\itemsep=1pt
\item $\mathtt{true}$,
\item $(\mathit{atrb}:v)$, i.e., an attribute-value pair,
\item $\phi\land\psi$,
\item $\phi\lor\psi$,
\item $\lnot\phi$.
\end{itemize}
If a sentence $\phi$ is \textit{satisfied} or \textit{true} in a state $s$, we write $s\Vdash\phi$. The semantics of $L$ is defined by
\begin{itemize}
\itemsep=1pt
\item $s\Vdash\mathtt{true}$ always,
\item $s\Vdash (\mathit{atrb}:v)\iff(\mathit{atrb}:v)\in s$,
\item $s\Vdash\phi\land\psi \iff s\Vdash\phi$ and $s\Vdash\psi$,
\item $s\Vdash\phi\lor\psi \iff s\Vdash\phi$ or $s\Vdash\psi$,
\item $s\Vdash\lnot\phi \iff$ not $s\Vdash\phi$.
\end{itemize}

Let $\Phi$ be a sentence in $L$.
When a sentence in $L$ appears in a written policy (see below), it is called a \textit{context}.

We define two kinds of plans: an \textit{attribute condition plan} is a triple $I:\Phi:\pi$, and a \textit{belief state condition plan} is a triple $I:B:\pi$, where $I$ is a set of intentions, $\pi$ is a POMDP policy, $\Phi$ is a context and $B$ is a belief state.
All plans are stored in a \textit{plan library}.

The idea is that attribute condition plans (abbreviation: a-plans) are written by agent designers and are available for use when the agent is deployed. Roughly speaking, belief state condition plans (abbreviation: b-plans) are automatically generated by a POMDP planner and stored when no a-plan is found which `matches' the agent's current belief state and intention set.

Policies in a-plans are of two kinds:
\begin{definition}[Most likely context]
An a-plan {\em most-likely-context} policy is either an action or has the form
\[
a:\{(\Phi_1,\pi_1),\ldots,(\Phi_n,\pi_n)\},
\]
where $a$ is an action, the $\Phi_i$ are contexts, and each of the $\pi_i$ is one of the two kinds of a-plan policies. 
\end{definition}
At belief state $B$, the \textit{degree of belief} of $\Phi$ is
\[
\mathit{Degree}(\Phi,B) :=\sum_{s\in S, s\Vdash \Phi}B(s).
\]

We abbreviate ``most-likely-context'' as `ml'.
If an ml policy $\pi=a:\mathit{ML}$ is adopted for execution and it is not simply an action, then $a$ is executed, an observation is received, the current belief state is updated to $B'$ and finally the policy which is paired with the most likely context is executed -- that is,
\[
\argmax_{\pi':\quad (\Phi',\pi')\in \mathit{ML}}\mathit{Degree}(\Phi',B')
\]
is executed.

\begin{definition}[First applicable context]
An a-plan {\em first-applicable-context} policy is either an action or has the form
\[
a:\langle(\Phi_1\bowtie p_1,\pi_1),\ldots,(\Phi_n\bowtie p_n,\pi_n)\rangle,
\]
where $a$ is an action, the $\Phi_i$ are contexts, $\bowtie := \{\leq,\geq\}$, the $p_i$ are probabilities, and each of the $\pi_i$ is one of the two kinds of a-plan policies. 
\end{definition}
We abbreviate ``first-applicable-context'' as `fa'.
If an fa policy $\pi=a:\mathit{FA}$ is adopted for execution and it is not simply an action, then $a$ is executed, an observation is received, the current belief state is updated to $B'$ and finally the policy which is paired with the \textit{first} context which satisfies its probability inequality is executed - that is, $\pi_i$ is executed such that $\mathit{Degree}(\Phi_i,B')\bowtie p_i$ and $(\Phi_i\bowtie p_i,\pi_i)\in \mathit{FA}$ and there is no $(\Phi_j\bowtie p_j,\pi_j)\in \mathit{FA}$ such that $j<i$ for which $\mathit{Degree}(\Phi_j,B')\bowtie p_j$. If no context in the sequence $\langle(\Phi_1\bowtie p_1,\pi_1),\ldots,(\Phi_n\bowtie p_n,\pi_n)\rangle$ satisfies its inequality, the a-plan of which the policy is a part is regarded as having finished, that is, the control loop is then in a position where a fresh plan in the plan library is sought.

In the following example a-plan policy, an agent must move around in a six-by-six grid world to collect items. Suppose the plan selected from the library is $I:\Phi:\pi$ with $I$ being
$\{\mathtt{six{-}one},\mathtt{collect}\}$,
$\Phi$ being
\[
((\mathtt{direction:North}) \lor (\mathtt{direction:West})) \land \lnot (\mathtt{x{-}coord}:6) \land \lnot (\mathtt{y{-}coord}:1)
\]
and $\pi$ being
\begin{align*}
& \mathtt{move{-}forward} : \{ \\
& \quad\quad((\mathtt{item{-}here}:\mathtt{yes}),\mathtt{take{-}item}),\\
& \quad\quad((\mathtt{item{-}here}:\mathtt{no}),\mathtt{move{-}forward} :  \langle\\
& \quad\quad\quad\quad((\mathtt{x{-}coord}:6),(\mathtt{direction:North})\geq 0.9,\mathtt{turn{-}left}),\\
& \quad\quad\quad\quad((\mathtt{y{-}coord}:1),(\mathtt{direction:West})\geq 0.9,\mathtt{turn{-}right}),\\
& \quad\quad\quad\quad(\mathtt{true}\leq 1),\mathtt{move{-}forward})\rangle\}	
\end{align*}
One can see that $\pi$ itself is an ml policy, but embedded inside it is an fa policy.

\begin{figure}[t]
\centering
\includegraphics[scale=1]{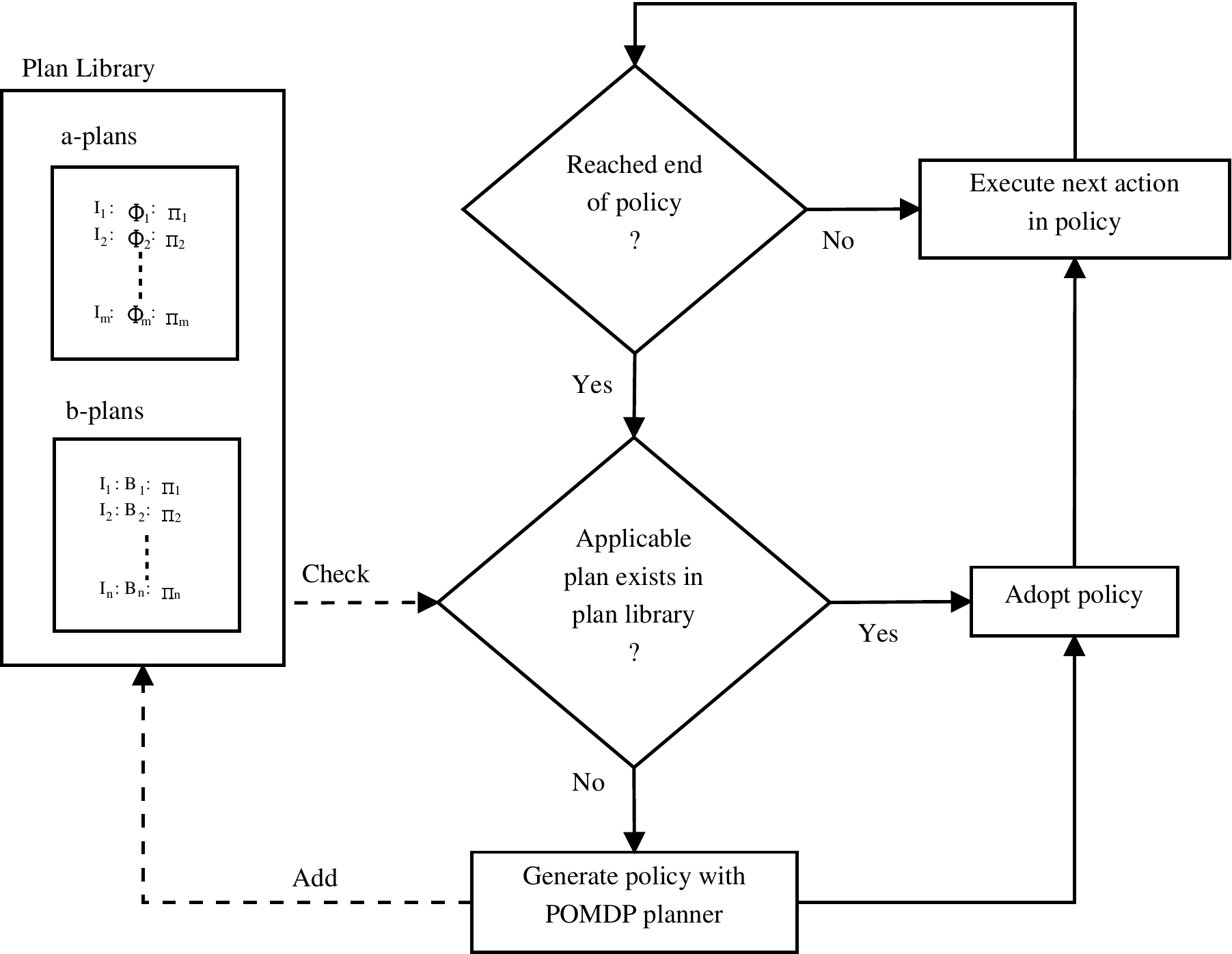}
\caption{A flow diagram of the planning process in the new version of the HPB agent architecture.\label{fig:planning}}
\end{figure}

Suppose that the agent currently has a belief state $B^\mathit{cur}$ and an intention set $I^\mathit{cur}$.
First, the agent will scan through all a-plans, selecting all those which `match' $I^\mathit{cur}$. From this set, the agent will execute the policy $\pi$ of the a-plan $I:\Phi:\pi$ whose attribute condition has the highest degree of belief at $B^\mathit{cur}$.
If the set of a-plans matching $I^\mathit{cur}$ is empty, the agent will scan through all b-plans, selecting all those which `match' $I^\mathit{cur}$. From this set, the agent will execute the policy $\pi$ of the b-plan $I:B:\pi$ whose belief state is `most similar' to $B^\mathit{cur}$.
If the set of b-plans matching $I^\mathit{cur}$ is empty, or there is no b-plan with belief state similar to $B^\mathit{cur}$, then the agent will generate policy $\pi^\mathit{cur}$, execute it and store $I^\mathit{cur}:B^\mathit{cur}:\pi^\mathit{cur}$ in the plan library for possible reuse later.
The high-level planning process is depicted by the diagram in Figure~\ref{fig:planning}.

To ``execute policy $\pi$'' (where $\pi$ has horizon/depth $h$) means to perform $h$ actions as recommended by $\pi$. No policy will be sought in the library, nor will a new policy be generated until the action recommendations of the current policy being executed have been `exhausted'. One may be concerned that a policy becomes `stale' or inapplicable while being executed, and that seeking or generating `fresh' policies at every iteration keeps action selection relevant in a dynamic world. However, written policies (in a-plans) should preferably have the form of generated policies, and generated policies (in b-plans) can deal with all situations understood by the agent: It is assumed that each observation distinguishable by the agent, identifies a particular state of the world, as far as the agent's sensors allow. Hence, if a policy considers every observation at its choice nodes, the policy will have a recommended (for written policies) or optimal (for generated policies) action, no matter the state of the world.
However, writing or generating policies with far horizons (e.g., $h>7$) is impractical. With large $h$, an agent will take relatively long to generate a policy and thus lose its reactiveness. Reactiveness is especially important in highly dynamic environments.

\begin{algorithm}[t]
\caption{$FindPolicy$}
\label{algo:2}
\KwIn{$B^\mathit{cur}$: current belief state} 
\KwIn{$I^\mathit{cur}$: current intention set}
\KwIn{$\theta_i$: intention-set threshold}
\KwIn{$\theta_b$: belief state threshold}
\KwIn{$\mathit{PlanLib}$: the plan library}
\KwIn{$h$: planning horizon / policy depth}
\KwOut{A POMDP policy of depth $h$}
$\mathit{ApplicablePlans} \gets \{I^{lib}:\Phi:\pi^{lib}\in\mathit{PlanLib} \mid \mathit{IS}(I^\mathit{cur},I^\mathit{lib})\geq \theta_i\}$\;
\If{$\mathit{ApplicablePlans}\neq\emptyset$}{
\Return $\argmax_{\pi^{lib}:\quad I^{lib}:\Phi:\pi^{lib}\in\mathit{ApplicablePlans}}\mathit{Degree}(\Phi,B^\mathit{cur})$\;}
\If{$\mathit{ApplicablePlans}=\emptyset$}{$\mathit{ApplicablePlans} \gets \{I^{lib}:B^{lib}:\pi^{lib}\in\mathit{PlanLib} \mid \mathit{BS}(I^\mathit{cur},I^\mathit{lib},B^{cur},B^{lib})\geq \theta_i\}$\;
}
\If{$\mathit{ApplicablePlans}\neq\emptyset$}{
$\mathit{ApplicablePlans} \gets \{I^{lib}:B^{lib}:\pi^{lib}\in\mathit{ApplicablePlans} \mid R'(B^{lib},B^{cur})\geq \theta_b\}$\;
}
\If{$\mathit{ApplicablePlans}\neq\emptyset$}{
\Return $\argmax_{\pi^{lib}:\quad I^{lib}:B^{lib}:\pi^{lib}\in\mathit{ApplicablePlans}} R'(B^{lib},B^{cur})$\;}
\If{$\mathit{ApplicablePlans}=\emptyset$}{
$\pi^{cur}\gets\mathit{Policy}(B^\mathit{cur},I^\mathit{cur},h)$\;
Add $I^{cur}:B^{cur}:\pi^{cur}$ to $\mathit{PlanLib}$\;
\Return $\pi^{cur}$\;
}
\end{algorithm}

With respect to a-plans, whether two intention sets match will be determined by how many goals they have in common. Thus, the similarity between $I^\mathit{cur}$ and $I^\mathit{lib}$ can be determined as follows.
\[
\mathit{IS}(I^\mathit{cur},I^\mathit{lib}):= \frac{\sum_{g\in G, g\in I^\mathit{cur}, g\in I^\mathit{lib}}1}{|I^\mathit{cur}\uni I^\mathit{lib}|}.
\]
$\mathit{IS}(\cdot)$ lies in $[0,1]$.
$I^\mathit{cur}$ and $I^\mathit{lib}$ need not have equal cardinality. Larger values of $\mathit{IS}(\cdot)$ mean more similarity / closer match. The agent designer can decide what value of $\mathit{IS}(I,I')$ constitutes a `match' between $I$ and $I'$ (see the discussion on ``thresholds'' below).

What constitutes a match between intention sets with respect to b-plans is different: Policies generated at two times $t$ and $t'$ might be significantly different for the same (similar) context(s) if the satisfaction levels of the intentions are significantly different at the two times. This is an important insight because policies of b-plans are \textit{generated}, not written.
Even though $\mathit{IS}(I^\mathit{cur},I^\mathit{lib})$ may constitute a `match' (with $I^\mathit{lib}$ in a b-plan), $\pi^\mathit{lib}$ might be completely impractical for pursuing $I^\mathit{cur}$.
The measure of similarity will be the sum of differences between satisfaction levels. Note that an intention's satisfaction levels can only be compared if the intention appears in both intention sets under consideration. We denote the similarity between two intention sets $I^\mathit{cur}$ and $I^\mathit{lib}$ as $\mathit{BS}(I^\mathit{cur},I^\mathit{lib},B^\mathit{cur},B^\mathit{lib})$ and define it as follows.
\[
\mathit{BS}(I^\mathit{cur},I^\mathit{lib},B^\mathit{cur},B^\mathit{lib}):= \frac{\sum_{g\in G, g\in I^\mathit{cur}, g\in I^\mathit{lib}}1-||\sigma^g(B^\mathit{cur})-\sigma^g(B^\mathit{lib})||}{|I^\mathit{cur}\uni I^\mathit{lib}|}.
\]
where $||x||$ denotes the absolute value of $x$.
$\mathit{BS}(\cdot)$ lies in $[0,1]$.
$I^\mathit{cur}$ and $I^\mathit{lib}$ need not have equal cardinality. Larger values of $\mathit{BS}(\cdot)$ mean more similarity / closer match. The agent designer can decide what value of $\mathit{BS}(I,I')$ constitutes a `match' between $I$ and $I'$.

For a fixed pair of intention sets, $\mathit{BS}(I^\mathit{cur},I^\mathit{lib},B^\mathit{cur},B^\mathit{lib}) \leq \mathit{IS}(I^\mathit{cur},I^\mathit{lib})$. That is, $\mathit{BS}(\cdot)$ is a stronger measure of similarity than $\mathit{IS}(\cdot)$. This is because with $\mathit{BS}(\cdot)$, intention satisfaction levels must also be similar. The stronger measure is required to filter out b-plans that seem similar when judged only on the commonality of their intentions, but not on their satisfaction levels. And there may be several b-plans in the library which would be judged similar by $\mathit{IS}(\cdot)$, but they have been added to the library exactly because they are indeed different when their satisfaction levels are taken into account. The following example should make this clear. Suppose that the following two b-plans are in the library: $\{g1,g4\}:B_1:\pi_1$ and $\{g1,g4\}:B_2:\pi_2$, where $B_1=\{(s_1,0.95),(s_2,0.05),(s_3,0),(s_4,0)\}$ and $B_2=\{(s_1,0),(s_2,0),(s_3,0.05),(s_4,0.95)\}$. And suppose $g1$ is most satisfied when the agent is in $s_1$, and $g4$ is most satisfied when the agent is in $s_4$.
A policy to pursue $\{g1,g4\}$ when starting in $B_1$ would rather suggest actions to move towards $s_4$, while a  policy to pursue $\{g1,g4\}$ when starting in $B_2$ would rather suggest actions to move towards $s_1$. The point is that although the two b-plans are identical with respect to the intention set, they have very different policies, due to their different belief states (and thus satisfaction levels).

We now prepare for the definition of similarity between two belief states.
The `directed divergence' \citep{k68,c75b} of belief state $C$ from belief state $B$ is defined as
\[
R(C,B) := \sum_{s\in S}C(s)\ln \frac{C(s)}{B(s)}.
\]
$R(C,B)$ is undefined when $B(s)=0$ while $C(s)>0$. When $C(s)=0$, then $R(C,B)=0$ because  $\lim_{x\to0}\ln(x)=0$.
Let
\[
\Pi(B) := \{C'\in\Pi \mid \forall s\in S, \mbox{ if }C'(s)>0,\mbox{ then }B(s)>0\},
\]
where $\Pi$ is the set of all probability distributions over the states $S$ (i.e., all belief states which can be induced from $S$).
That is, $\Pi(B)$ is the set of belief states which keep $R(C',B)$ defined.
Let
\[
\mathit{maxR}(B) := \max_{C'\in \Pi(B)}R(C',B),
\]
For our purposes, we can define $R(C,B)$ as $\mathit{maxR}(C,B)$ whenever it would normally be undefined.
We define a slightly modified cross-entropy $R'$ as
\begin{equation*}
R'(C,B) := \left\{
\begin{array}{rl}
R(C,B) & \text{if }R(C,B)\text{ is defined}\\
\mathit{maxR}(B) & \text{otherwise}
\end{array} \right.
\end{equation*}
Finally, the similarity between the current belief state $B^{cur}$ and the belief state $B^{lib}$ in a plan in the library is
$R'(B^{lib},B^{cur})$.

Two thresholds are involved with determining when library plans are applicable and how plans are dealt with: the \textit{intention-set threshold} (abbreviation: $\theta_i$) and the \textit{belief-state threshold} (abbreviation: $\theta_b$). The former is involved in both a-plans and b-plans, and the latter is involved only in b-plans.

The $\mathit{FindPolicy}$ procedure (Algo.~\ref{algo:2}) formally defines what policy the agent will execute whenever the agent seeks a policy, and the procedure defines when and how new plans are added to the plan library.

\section{Simulations}
\label{sec:Simulations}

We performed some tests on an HPB agent in two domains: a six-by-six grid-world and a three-battery system.
In the experiments which follow, the threshold $\theta$ is set to $0.05$, $\mathit{MRY}$ is set to 5 and $h=3$. Desire levels are initially set to zero for all goals.
For each experiment, 10 trials were run.
The plan library is not made use of.
 
In the grid-world, the agent's task is to visit each of the four corners, and to collect twelve items randomly scattered.
The goals are $\{(1,1)$, $(1,6)$, $(6,1)$, $(6,6)$, $\mathtt{collect}\}$, and $(1,1)$, $(1,6)$, $(6,1)$ and $(6,6)$ are marked mutually incompatible. That is,
\begin{itemize}
\item $\mathit{Cpbl}((1,1))=\{(1,1),\mathtt{collect}\}$,
\item $\mathit{Cpbl}((1,6))=\{(1,6),\mathtt{collect}\}$,
\item $\mathit{Cpbl}((6,1))=\{(6,1),\mathtt{collect}\}$,
\item $\mathit{Cpbl}((6,6))=\{(6,6),\mathtt{collect}\}$,
\item $\mathit{Cpbl}(\mathtt{collect})=\{\mathtt{collect},(1,1),(1,6),(6,1),(6,6)\}$.
\end{itemize}
States are quadruples $\langle x,y,d,i\rangle$, with $x,y\in\{1,\cdots,6\}$ being the coordinates of the agent's position in the world, $d\in\{\mathtt{North},\mathtt{East},\mathtt{West},\mathtt{South}\}$ the direction it is facing, and $i\in\{0,1\}$, $i=1$ if an item is present in the cell with the agent, else $i=0$. The agent can perform five actions $\{\mathtt{left},\mathtt{right},\mathtt{forward},\mathtt{see},\mathtt{take}\}$, meaning, turn left, turn right, move one cell forward, see whether an item is present and take an item. The only observation possible when executing one of the physical actions is $\mathtt{obsNil}$, the null observation, and $\mathtt{see}$ has possible observations from the set $\{0,1\}$ for whether the agent sees the presence of an item (1) or not (0).

Next, we define the possible outcomes for each action: When the agent turns left or right, it can get stuck in the same direction, turn $90^\circ$ or overshoots by $90^\circ$. When the agent moves forward, it moves one cell in the direction it is facing or it gets stuck and does not move. The agent can see an item or see nothing (no item in the cell), and taking is deterministic (if there is an item present, it will be collected with certainty, if the agent executes $\mathtt{take}$). All actions except $\mathtt{take}$ are designed so that the correct outcome is achieved $95\%$ of the time and incorrect outcomes are achieved $5\%$ of the time.



Seven experiments were performed, with different weight-combinations ($\alpha_g$) assigned for each experiment.
For each trial, the agent starts in a random location and performs 100 actions. When $g\in\{(1,1),(1,6),(6,1),(6,6)\}$, we let $\sigma^g(s) = 1 - \mathit{dist}/10$ where 10 is the maximum Manhattan distance between two cells in the world and $\mathit{dist}$ is the Manhattan distance between the cells represented by $g$ and $s$, and we let
$\sigma^\mathtt{collect}(s) = 1 - \mathit{dist}/10$, where $\mathit{dist}$ is the Manhattan distance between the cell represented by $s$ and the closest cell containing an item.

In the BatryPack domain, the agent's task is to keep a pack of rechargeable batteries within a given voltage range. 
There are three batteries available, each with a maximum capacity of 6 volts. For every time-unit that a battery is in the pack, it loses 1 volt. For every time-unit a battery is out of the pack, it gains 1 volt. The pack is considered to be within range if the sum of the batteries currently in the pack is in $[3,9]$. (The possible pack-voltage-range is $[0,18]$).
The goals are $\{\mathtt{maintain},\mathtt{charge}\}$, meaning that the agent should, respectively, try to keep the pack within range, and charge the batteries. The goals are, intuitively, not mutually exclusive.
States are of the form $\langle (io_1,v_1),(io_2,v_2),(io_3,v_3)\rangle$, with $io_j$ being either $\mathtt{in}$ or $\mathtt{out}$, indicating whether battery $j$ is in or out of the pack, and $v_j$ is the batteries current voltage.
The agent can perform ten actions: $\mathtt{no{-}op}$ and for $j=1,2,3$, $\mathtt{remove}_j$, $\mathtt{add}_j$ and $\mathtt{measure}_j$, meaning, remove battery $j$ from the pack, add battery $j$ to the pack, respectively, measure the current voltage available in battery $j$. The only observation possible when executing one of the physical actions is $\mathtt{obsNil}$; $\mathtt{measure}_j$ has possible observations from the set $\{0,1,2,3,4,5,6\}$.
When the agent removes or adds a battery, it may fail to do so with a $5\%$ chance.
The measurement action is deterministic, but $5\%$ of the time it will perceive a voltage one volt more or less than it is actually.

Six experiments were performed with different weights ($W(g)$) assigned for each experiment - three while the goals were mutually incompatible and three while the goals may be pursued simultaneously.
For each trial, the initial state is $\langle (\mathtt{in},6),(\mathtt{out},3),(\mathtt{out},3)\rangle$ with the initial intention being $\mathtt{maintain}$ - and the agent/system performs 50 actions.
We let $\sigma^\mathtt{maintain}(s)$ equal 1 if the battery pack is within range, else it equals a value less than 1 (min. 0) in proportion to how far the pack voltage is from being within range.
We let
$\sigma^\mathtt{charge}(s)=c/3$ where $c$ is the number of batteries which are out of the pack.

\subsection{Evaluation of HPB Agent Performance}
\label{Evaluation}

Table~\ref{tbl:grid-world-agent} shows the results. It can be seen quite clearly that the agent can be directed to certain corners and to collect items with a dedication proportional to the weights chosen by the agent designer for the respective goals.

\begin{table}[t]
\caption{Performance of the grid-world agent for various combinations of goal-weights.
}
\centering
\begin{tabular}{| >{\centering\arraybackslash}m{19mm} | >{\centering\arraybackslash}m{6mm} | >{\centering\arraybackslash}m{6mm} | >{\centering\arraybackslash}m{6mm} | >{\centering\arraybackslash}m{6mm} | >{\centering\arraybackslash}m{6mm} | >{\centering\arraybackslash}m{7mm}| >{\centering\arraybackslash}m{6mm}|}
\hline
Goal	& \multicolumn{7}{|c|}{Weights}\\
\hline
(1,1) 				& 0 & 0 & 0.5 	& 0.5 	& 0 	& 0.33 	& 0.2 \\
(1,6) 				& 0 & 0 & 0.5 	& 0 	& 0		& 0.33 	& 0.2 \\
(6,6) 				& 0 & 1 & 0		& 0.5 	& 0.5 	& 0 	& 0.2 \\
(6,1) 				& 0 & 0 & 0 	& 0 	& 0 	& 0 	& 0.2 \\
$\mathtt{collect}$	& 1 & 0 & 0 	& 0 	& 0.5 	& 0.34 	& 0.2 \\
\hline
Goal	& \multicolumn{7}{|c|}{Performance}\\
\hline
(1,1) 				& 0.5 	& 0 	& 4.1 	& 4.1 	& 0 	& 3.5 	& 2.3 \\
(1,6) 				& 0.4 	& 0.1 	& 4.2 	& 0 	& 0.9	& 3.5 	& 2.2 \\
(6,6) 				& 0.3 	& 1		& 0		& 4.1 	& 4.6 	& 0.1 	& 2.3 \\
(6,1) 				& 0.5 	& 0 	& 0 	& 0 	& 0.4 	& 0.1 	& 2.5 \\
$\mathtt{collect}$ 	& 11 	& 0.2 	& 0.3 	& 0.4 	& 3.5 	& 1.6 	& 0.6 \\
\hline
\end{tabular}
\label{tbl:grid-world-agent}
\end{table}

Table~\ref{tbl:batry-pack} shows the results. It can be seen that the system performs better when the goals are pursued jointly, particularly when maintaining pack voltage and trying to charge the batteries have equal weights. In the table, ``$\%$ one/two intentions'' means that for $x/y$, the percentage of time that there was exactly one intention in $I$ is $x$ and the percentage of time that there were exactly two intentions in $I$ is $y$.

\begin{table}[t]
\caption{Performance of the BatryPack system for various combinations of goal-weights, for mutually compatible (joint) and incompatible (disjoint) goals.
}
\centering
\begin{tabular}{| >{\centering\arraybackslash}m{30mm} | >{\centering\arraybackslash}m{11mm} | >{\centering\arraybackslash}m{11mm} | >{\centering\arraybackslash}m{11mm} |}
\hline
\multicolumn{4}{|c|}{Joint}\\
\hline
$\mathtt{maintain}$ weight	& 0.2	& 0.5	& 0.8\\
$\mathtt{charge}$ weight	& 0.8	& 0.5	& 0.2\\
$\%$ within range				& 50	& 65.8	& 56.6\\
$\%$ one/two intentions			& 19/81	& 18/82	& 58/42\\
\hline
\end{tabular}
\begin{tabular}{| >{\centering\arraybackslash}m{30mm} | >{\centering\arraybackslash}m{11mm} | >{\centering\arraybackslash}m{11mm} | >{\centering\arraybackslash}m{11mm} |}
\hline
\multicolumn{4}{|c|}{Disjoint}\\
\hline
$\mathtt{maintain}$ weight	& 0.2	& 0.5	& 0.8\\
$\mathtt{charge}$ weight	& 0.8	& 0.5	& 0.2\\
$\%$ within range				& 52	& 53.6	& 48.8\\
$\%$ one/two intentions			& 100/0	& 100/0	& 100/0\\
\hline
\end{tabular}
\label{tbl:batry-pack}
\end{table}

These experiments highlight four important features of the HPB architecture:\\
(1) Each of several goals can be pursued individually until satisfactorily achieved. (2) Goals must periodically be re-achieved. (3) The trade-off between (weights of) goals can be set effectively. (4) Goals can be satisfied even while dealing with stochastic actions and perceptions.

\section{Related Work}
\label{Related-Work}

AgentSpeak$^+$ \citep{bmhclgs} extends the BDI language AgentSpeak \citep{r96} with on-demand probabilistic planning in uncertain environments. AgentSpeak has a plan library of plans, each plan being of the form
\[
e:b_1\land \cdots \land b_m \gets c_1;\cdots ; c_n
\]
where $e$ is a \textit{triggering event}, $b_1, \ldots, b_m$ are belief literals and $c_1,\ldots, c_n$ are actions or goals.
Goals may become (internal) triggering events. Events in the (external) environment may also be perceived as triggering events. As triggering events occur, they are placed in a set and periodically selected for processing. An event is `processed' by selecting an appropriate plan from the plan library with a matching triggering event. A plan is appropriate if its \textit{context} $b_1\land \cdots \land b_m$ is a logical consequence of the agent's set of \textit{base beliefs}.
The goals and/or actions $c_1;\cdots ; c_n$ of the selected appropriate plan will be processed in sequence. If $c_i$ is an action, it is executed; if it is a goal, it becomes an internal event which may trigger the selection and execution of further plans.
An AgentSpeak agent maintains a set of intentions and each intention is a stack of plans. Please refer to \citep{r96} for details.
When considering HPB plans, $e$ is roughly analogous to $I$, $b_1\land \cdots \land b_m$ is roughly analogous to $\Phi$ or $B$ and $c_1;\cdots ; c_n$ is roughly analogous to $\pi$.

The contribution of AgentSpeak$^+$ is to allow a POMDP planner to suggest the optimal action at a point in a (written) plan where the agent designer feels that an \textit{optimal} action is required at that point, or that there is insufficient information at the time of writing the plan to suggest a reasonable action. In other words, there might be points in a plan when actions are best chosen just before execution so that they can be determined appropriately for the agent's \textit{current} context.

\citet{bmhclgs} make use of only the first action of any POMDP policy.
Online POMDP planners do forward-search to a given depth $h$ (number of future actions). The deeper the look-ahead depth, the more optimal the actions in the policy. It might actually be a waste of computational resources to discard the whole policy of depth $h$ once it is available. An agent could use its whole policy-tree and only generate a new policy after it has finished using the current policy to execute $h$ actions. However, the actions closer to the end of the policy tree will tend to be farther from optimal than those closer to the tree's root. In future work, we would like to find ways to balance out the myopic take-first-action approach and the over-optimistic take-all-actions approach.

AgentSpeak$^+$ does not have a mechanism for storing and reusing generated policies.

An advantage of AgentSpeak$^+$ is that their written plans can be more expressive than HPB plans: elements of their plans are written in a language based on a fragment of first-order logic, including n-ary predicates and variable terms.
Nonetheless, even though an HPB a-plan is propositional in nature (not relational), a policy has a reasonably expressive tree structure with branching conditional on observations of context sentences. A desirable feature that AgentSpeak plans have that HPB plans lack is the ability to call plans from within plans.

Some slightly less related work will now be reviewed.

\citet{wbpl07} and \citet{mzml07} have incorporated online plan generation into BDI systems, however the planners deal only with deterministic actions and observations.

\citet{nt05} use POMDP theory to coordinate teams of agents.  However, their framework is very different to our architecture. They use POMDP theory to determine good role assignments of team members, not for generating policies online.

\citet{ldap08} provide a rather sophisticated architecture for controlling the behavior of an emotional agent. Their agents reason with several classes of emotion and their agents are supposed to portray emotional behavior, not simply to solve problems, but to look believable to humans. Their architecture has a ``continuous planner [...] that is capable of partial order planning and includes emotion-focused coping [...]'' Their work has a different application to ours, however, we could take inspiration from them to improve the HPB architecture.

\citet{pgdc08} take a different approach to use POMDPs to improve BDI agents. By leveraging the relationship between POMDP and BDI models, as discussed by \citet{sp06}, they devised an algorithm to extract BDI plans from optimal POMDP policies. The main difference to our work is that their policies are pre-generated and BDI-style rules are extracted for all contingencies. The advantage is that no (time-consuming) online plan/policy generation is necessary. The disadvantage of their approach is that all the BDI plans must be stores and every time the domain model changes, a new POMDP must be solved and the policy-to-BDI-plan algorithm must be run. It is not exactly clear from their paper \citep{pgdc08} how or when intentions are chosen.
Although it is interesting to know the relationship between POMDPs and BDI models \citep{sp06,sp11}, we did not use any of these insights in developing our architecture. However, the fact that the HPB architecture does integrate the two frameworks, is probably due to the existence of the relationship.

\citet{rfv09} also introduced a hybrid POMDP-BDI architecture, but without a notion of desire levels or satisfaction levels. Although their basic approaches to combine the POMDP and BDI frameworks is the same as ours, there are at least three major differences: Firstly, they define their architecture in terms of the GOLOG agent language \citep{brst00}. Secondly, their approach uses a computationally intensive method for deciding whether to refocus; performing short policy look-aheads to ascertain the most valuable goal to pursue.\footnote{Essentially, the goals in $G$ are stacked in descending order of the value of $V^*_\mathit{HPB}(B,g,h^-)$, where $h^- < h$ and $B$ is the current belief state. The goal on top of the stack becomes the intention.} Our approach seems much more efficient. Thirdly, in their approach, the agent cannot pursue several goals concurrently.

\citet{chlgsl13} incorporate probabilistic graphical models into the BDI framework for plan selection in stochastic environments.
An agent maintains epistemic states (with random variables) to model the uncertainty about the stochastic environment, and corresponding belief sets of the epistemic state are defined. The possible states of the environment, according to sensory observations, and their relationships are modeled using probabilistic graphical models: The uncertainty propagation is carried out by Bayesian Networks and
belief sets derived from the epistemic states trigger the selection of relevant
plans from a plan library. For cases when more than one plan is applicable due to uncertainty in an agent's beliefs, they propose a utility-driven
approach for plan selection, where utilities of actions are modeled in influence diagrams.
Our architecture is different in that it does not have a library of pre-supplied plans; in our architecture, policies (plans) are generated online.

None of the approaches mentioned maintain desire levels for selecting intentions. The benefit of maintaining desire levels is that intentions are not selected only according what they offer with respect to their \textit{current} expected reward, but also according to when last they were achieved.

\section{Conclusion}
\label{Conclusion}

Our work focuses on providing high-level decision-making capabilities for robots and agents who live in dynamic stochastic environments, where multiple goals and goal types must be pursued.
We introduced a hybrid POMDP-BDI agent architecture, which may display emergent behavior, driven by the intensities of their desires. In the past decade, several BDIAs have been augmented with capabilities to deal with uncertainty. The HPB architecture is novel in that it can pursue multiple goals concurrently. Goals must periodically be re-achieved, depending on the goals' desire levels, which change over time and in proportion to how close the goals are to being satisfied.

A major benefit of the HPB architecture is that 
every action recommended by a generated policy simultaneously maximizes the agent's reward with respect to pursuit of \textit{all} the current intentions. As far as the authors are aware, no other agent architecture is capable of this.

In previous work \citep{rm15a}, we argued that maintenance goals like avoiding moist areas (or collecting soil samples) should rather be viewed as a \textit{preference} and modeled as a POMDP reward function. And specific tasks to complete (like collecting gas or keeping its battery charged) should be modeled as BDI desires. The idea is that while the agent is pursuing goals, it can concurrently perform rewarding actions not directly related to its goals. The architecture reported about in this paper does not make a clear distinction between overt and maintenance goals. In the new version of the architecture, that distinction can be simulated, however, now goals can be pursued in a much more fine-grained way via the choice of goal-weights ($W(g)$).

Another important feature brought into the new version is the ability to mark sets of goals as \textit{disjoint} thereby forcing the agent to never pursue these goals concurrently, that is, disjoint goals will never be in $I$ simultaneously.

Although \citet{nt05} and \citet{chlgsl13} call their approaches hybrid, our architecture can arguably more confidently be called hybrid because of its more intimate integration of POMDP and BDI concepts.

We could take some advice from \citet{ap11}. They provide a systematic methodology to incorporate emotion into a decision-theoretic framework, and also provide ``a principled, domain-independent methodology for generating heuristics in novel situations''.

Policies returned by $\mathit{Plan}$ as defined in this paper are optimal. A major benefit of a POMDP-based architecture is that the literature on POMDP planning optimization \citep{m00,rgt05,ptc05,lcl05,sbs07,rppc08,clc09,spk13} (for instance)
can be drawn upon to improve the speed with which policies can be generated.

Evaluating the proposed architecture in richer domains would highlight problems in the architecture and indicate new directions for research and development in the area of hybrid POMDP-BDI architectures.

The expressivity of the language we use for describing goals and for writing conditions in a-plans is relatively low. AgentSpeak, for instance, has a richer language. The language's expressivity is mostly independent of the architecture. We thus chose to use a simple language to better focus on the components we want to discuss.

The design of the HPB agent architecture is a medium-to-long-term programme. We would like to keep improving its capabilities to deal with unforeseen, complex events in a changing, noisy environment.
The next step is to rigorously test the architecture using an HPB agent in a complex simulated world.
In particular, HPB agents with a plan library, including (pre-written) a-plans and (generated) b-plans, must still be assessed.
There is also scope for improving the focussing procedure. And analyzing under what conditions the two forms of desire update rule produce better performance must be investigated.

There may be better methods for learning than policy reuse. Policy reuse has its place when reasoning time or power is limited, but given the time and power, more sophisticated techniques could perhaps generate and store shorter, more effective plans. For instance, when an agent encounters a landmark with relatively high certainty, the landmark's location can be stored. The agent could then augment its sensor readings with the stored location data to reach the landmark more easily in future. Some objects in the environment might not be stable, and their location data should `degrade' over time in proportion to the environment's dynamism.

\citet{sspj11} provide a method for learning which (pre-written) plans in a BDI system should be executed in which contexts (given a selection of context-applicable plans). Their approach can also relearn context-plan matches as conditions change in dynamic environments. Future versions of the HPB architecture could benefit from ideas in their work.

Prediction is an inherent part of POMDP planning, but we would like our agents to predict much farther into the future, and recognize critical events which it should deal with or avoid. POMDP policies and pre-written plans are more for local `tactical' control. We need to bring in techniques for the agent to think globally or `strategically'.

The set of intentions might change while executing a policy. If the current set of intentions changes a lot, the current policy might become inapplicable. This is a typical BDI \textit{reconsideration} issue. However, an HPB agent will usually only perform very few actions before seeking a new plan. Just as in the case with humans, our agent should normally not get in trouble by assuming that things have not changed significantly in the last few steps. If the environment is so dynamic that relatively short plans can become inappropriate before completion of the plans, then the agent should have some more low-level, reactive systems to deal with the changes. In highly dynamical environments, the HPB `agent' is better suited to being the high-level reasoning module of a larger system.


\bibliographystyle{apalike}
\bibliography{references}

\begin{thebibliography}{}

\bibitem[Antos and Pfeffer, 2011]{ap11}
Antos, D. and Pfeffer, A. (2011).
\newblock Using emotions to enhance decision-making.
\newblock In Walsh, T., editor, {\em Proceedings of the Twenty-second Intl.
  Joint Conf. on Artif. Intell. (IJCAI-11)}, pages 24--30, Menlo Park, CA. AAAI
  Press.

\bibitem[Bauters et~al., 2015]{bmhclgs}
Bauters, K., McAreavey, K., Hong, J., Chen, Y., Liu, W., Godo, L., and Sierra,
  C. (2015).
\newblock Probabilistic planning in agentspeak using the pomdp framework.
\newblock In Hatzilygeroudis, I., Palade, V., and Prentzas, J., editors, {\em
  Combinations of Intelligent Methods and Applications: Proceedings of the
  Fourth Intl. Workshop, CIMA 2014}, volume~46 of {\em Smart Innovation,
  Systems and Technologies}. Springer.

\bibitem[Boutilier et~al., 2000]{brst00}
Boutilier, C., Reiter, R., Soutchanski, M., and Thrun, S. (2000).
\newblock Decision-theoretic, high-level agent programming in the situation
  calculus.
\newblock In {\em Proceedings of the Seventeenth Natl. Conf. on Artif. Intell.
  (AAAI-00) and of the Twelfth Conf. on Innovative Applications of Artif.
  Intell. (IAAI-00)}, pages 355--362. AAAI Press, Menlo Park, CA.

\bibitem[Bratman, 1987]{b87}
Bratman, M. (1987).
\newblock {\em Intention, Plans, and Practical Reason}.
\newblock Harvard University Press, Massachusetts/England.

\bibitem[Cai et~al., 2009]{clc09}
Cai, C., Liao, X., and Carin, L. (2009).
\newblock Learning to explore and exploit in pomdps.
\newblock In {\em NIPS}, pages 198--206.

\bibitem[Chen et~al., 2013]{chlgsl13}
Chen, Y., Hong, J., Liu, W., Godo, L., Sierra, C., and Loughlin, M. (2013).
\newblock Incorporating {PGMs} into a {BDI} architecture.
\newblock In Boella, G., Elkind, E., Savarimuthu, B., Dignum, F., and Purvis,
  M., editors, {\em PRIMA 2013: Principles and Practice of Multi-Agent
  Systems}, volume 8291 of {\em Lecture Notes in Computer Science}, pages
  54--69. Springer, Berlin/Heidelberg.

\bibitem[Csisz\'ar, 1975]{c75b}
Csisz\'ar, I. (1975).
\newblock I-divergence geometry of probability distributions and minimization
  problems.
\newblock {\em Annals of Probability}, 3:146--158.

\bibitem[Kaelbling et~al., 1998]{klc98}
Kaelbling, L., Littman, M., and Cassandra, A. (1998).
\newblock Planning and acting in partially observable stochastic domains.
\newblock {\em Artif. Intell.}, 101(1--2):99--134.

\bibitem[Kinny and Georgeff, 1991]{kg91}
Kinny, D. and Georgeff, M. (1991).
\newblock Commitment and effectiveness of situated agents.
\newblock In {\em Proceedings of the 12th Intl. Joint Conf. on Artif. Intell.
  (IJCAI-91)}, pages 82--88.

\bibitem[Kinny and Georgeff, 1992]{kg92}
Kinny, D. and Georgeff, M. (1992).
\newblock Experiments in optimal sensing for situated agents.
\newblock In {\em Proceedings of the the Second Pacific Rim Intl. Conf. on
  Artif. Intell. (PRICAI-92)}.

\bibitem[Koenig, 2001]{k01}
Koenig, S. (2001).
\newblock Agent-centered search.
\newblock {\em Artif. Intell. Magazine}, 22:109--131.

\bibitem[Kullback, 1968]{k68}
Kullback, S. (1968).
\newblock {\em Information theory and statistics}, volume~1.
\newblock Dover, New York, 2nd edition.

\bibitem[Li et~al., 2005]{lcl05}
Li, X., Cheung, W., and Liu, J. (2005).
\newblock Towards solving large-scale {POMDP} problems via spatio-temporal
  belief state clustering.
\newblock In {\em Proceedings of IJCAI-05 Workshop on Reasoning with
  Uncertainty in Robotics (RUR-05)}.

\bibitem[Lim et~al., 2008]{ldap08}
Lim, M., Dias, J., Aylett, R., and Paiva, A. (2008).
\newblock Improving adaptiveness in autonomous characters.
\newblock In Prendinger, H., Lester, J., and Ishizuka, M., editors, {\em
  Intelligent Virtual Agents}, volume 5208 of {\em Lecture Notes in Computer
  Science}, pages 348--355. Springer, Berlin/Heidelberg.

\bibitem[Lovejoy, 1991]{l91}
Lovejoy, W. (1991).
\newblock A survey of algorithmic methods for partially observed {M}arkov
  decision processes.
\newblock {\em Annals of Operations Research}, 28:47--66.

\bibitem[Meneguzzi et~al., 2007]{mzml07}
Meneguzzi, F., Zorzo, A., M\'ora, M., and M., L. (2007).
\newblock Incorporating planning into {BDI} systems.
\newblock {\em Scalable Computing: Practice and Experience}, 8(1):15--28.

\bibitem[Monahan, 1982]{m82}
Monahan, G. (1982).
\newblock A survey of partially observable {M}arkov decision processes: Theory,
  models, and algorithms.
\newblock {\em Management Science}, 28(1):1--16.

\bibitem[Murphy, 2000]{m00}
Murphy, R. (2000).
\newblock {\em Introduction to AI Robotics}.
\newblock MIT Press, Massachusetts/England.

\bibitem[Nair and Tambe, 2005]{nt05}
Nair, R. and Tambe, M. (2005).
\newblock Hybrid bdi-pomdp framework for multiagent teaming.
\newblock {\em J. Artif. Intell. Res.(JAIR)}, 23:367--420.

\bibitem[Paquet et~al., 2005]{ptc05}
Paquet, S., Tobin, L., and Chaib-draa, B. (2005).
\newblock Real-time decision making for large {POMDPs}.
\newblock In {\em Advances in Artif. Intell.: Proceedings of the Eighteenth
  Conf. of the Canadian Society for Computational Studies of Intelligence},
  volume 3501 of {\em Lecture Notes in Computer Science}, pages 450--455.
  Springer Verlag.

\bibitem[Pereira et~al., 2008]{pgdc08}
Pereira, D., Gon\c{c}alves, L., Dimuro, G., and Costa, A. (2008).
\newblock Constructing bdi plans from optimal pomdp policies, with an
  application to agentspeak programming.
\newblock In G.~Henning, M.~G. and Goneet, S., editors, {\em XXXIV
  Confer\^encia Latinoamericano de Inform\'atica, Santa Fe. Anales CLEI 2008},
  pages 240--249.

\bibitem[Pollack and Ringuette, 1990]{pr90}
Pollack, M. and Ringuette, M. (1990).
\newblock Introducing the {T}ileworld: {E}xperimentally evaluating agent
  architectures.
\newblock In {\em Proceedings of the Eighth Conf. on Artif. Intell.}, pages
  183--189. AAAI Press.

\bibitem[Rao, 1996]{r96}
Rao, A. (1996).
\newblock {A}gent{S}peak({L}): {BDI} agents speak out in a logical computable
  language.
\newblock In {\em Proceedings of the 7th European Workshop on Modelling
  Autonomous Agents in a Multi-Agent World (MAAMAW-96)}, pages 42--55,
  Berlin/Heidelberg. Springer Verlaag.

\bibitem[Rao and Georgeff, 1995]{rg95}
Rao, A. and Georgeff, M. (1995).
\newblock {BDI} agents: {F}rom theory to practice.
\newblock In {\em Proceedings of the ICMAS-95}, pages 312--319. AAAI Press.

\bibitem[Rens et~al., 2009]{rfv09}
Rens, G., Ferrein, A., and {Van~der~Poel}, E. (2009).
\newblock A {BDI} agent architecture for a {POMDP} planner.
\newblock In Lakemeyer, G., Morgenstern, L., and Williams, M.-A., editors, {\em
  Proceedings of the Ninth Intl. Symposium on Logical Formalizations of
  Commonsense Reasoning (Commonsense 2009)}, pages 109--114, University of
  Technology, Sydney. UTSe Press.

\bibitem[Rens and Meyer, 2015]{rm15a}
Rens, G. and Meyer, T. (2015).
\newblock Hybrid {POMDP-BDI}: An agent architecture with online stochastic
  planning and desires with changing intensity levels.
\newblock In Duval, B., Van~den Herik, J., Loiseau, S., and Filipe, J.,
  editors, {\em Proceedings of the Seventh Intl. Conf. on Agents and Artif.
  Intell. (ICAART), Revised Selected Papers}, LNAI, pages 79--99. Springer
  Verlaag.

\bibitem[Ross et~al., 2008]{rppc08}
Ross, S., Pineau, J., Paquet, S., and Chaib-draa, B. (2008).
\newblock Online planning algorithms for {POMDPs}.
\newblock {\em Journal of Artif. Intell. Research (JAIR)}, 32:663--704.

\bibitem[Roy et~al., 2005]{rgt05}
Roy, N., Gordon, G., and Thrun, S. (2005).
\newblock Finding approximate {POMDP} solutions through belief compressions.
\newblock {\em Journal of Artif. Intell. Research (JAIR)}, 23:1--40.

\bibitem[Schut and Wooldridge, 2000]{sw00}
Schut, M. and Wooldridge, M. (2000).
\newblock Intention reconsideration in complex environments.
\newblock In {\em Proceedings of the the Fourth Intl. Conf. on Autonomous
  Agents (AGENTS-00)}, pages 209--216, New York, NY, USA. ACM.

\bibitem[Schut and Wooldridge, 2001]{sw01b}
Schut, M. and Wooldridge, M. (2001).
\newblock The control of reasoning in resource-bounded agents.
\newblock {\em The Knowledge Engineering Review}, 16(3):215--240.

\bibitem[Schut et~al., 2004]{swp04}
Schut, M., Wooldridge, M., and Parsons, S. (2004).
\newblock The theory and practice of intention reconsideration.
\newblock {\em Experimental and Theoretical Artif. Intell.}, 16(4):261--293.

\bibitem[Shani et~al., 2007]{sbs07}
Shani, G., Brafman, R., and Shimony, S. (2007).
\newblock Forward search value iteration for {POMDPs}.
\newblock In de~Mantaras, R.~L., editor, {\em Proceedings of the Twentieth
  Intl. Joint Conf. on Artif. Intell. (IJCAI-07)}, pages 2619--2624, Menlo
  Park, CA. AAAI Press.

\bibitem[Shani et~al., 2013]{spk13}
Shani, G., Pineau, J., and Kaplow, R. (2013).
\newblock A survey of point-based pomdp solvers.
\newblock {\em Autonomous Agents and Multi-Agent Systems}, 27(1):1--51.

\bibitem[Simari and Parsons, 2006]{sp06}
Simari, G. and Parsons, S. (2006).
\newblock On the relationship between mdps and the bdi architecture.
\newblock In {\em Proceedings of the Fifth Intl. Joint Conf. on Autonomous
  Agents and Multiagent Systems}, AAMAS '06, pages 1041--1048, New York, NY,
  USA. ACM.

\bibitem[Simari and Parsons, 2011]{sp11}
Simari, G. and Parsons, S. (2011).
\newblock {\em Markov Decision Processes and the Belief-Desire-Intention
  Model}.
\newblock Springer Briefs in Computer Science. Springer, New York, Dordrecht,
  Heidelberg, London.

\bibitem[Singh et~al., 2011]{sspj11}
Singh, D., Sardina, S., Padgham, L., and James, G. (2011).
\newblock Integrating learning into a {BDI} agent for environments with
  changing dynamics.
\newblock In Walsh, T., editor, {\em Proceedings of the Twenty-Second Intl.
  Joint Conf. on Artif. Intell. (IJCAI-11)}, pages 2525--2530, Menlo Park, CA.
  AAAI Press.

\bibitem[Walczak et~al., 2007]{wbpl07}
Walczak, A., Braubach, L., Pokahr, A., and Lamersdorf, W. (2007).
\newblock Augmenting {BDI} agents with deliberative planning techniques.
\newblock In Bordini, R., Dastani, M., Dix, J., and Seghrouchni, A., editors,
  {\em Proceedings of the Fourth Intl. Workshop of Programming Multi-Agent
  Systems (ProMAS-06)}, pages 113--127, Heidelberg/Berlin. Springer Verlag.

\bibitem[Wooldridge, 1999]{w99}
Wooldridge, M. (1999).
\newblock Intelligent agents.
\newblock In Weiss, G., editor, {\em Multiagent Systems: {A} Modern Approach to
  Distributed Artif. Intell.}, chapter~1. MIT Press, Massachusetts/England.

\bibitem[Wooldridge, 2000]{w00}
Wooldridge, M. (2000).
\newblock {\em Reasoning about Rational Agents}.
\newblock MIT Press, Massachusetts/England.

\bibitem[Wooldridge, 2002]{w02}
Wooldridge, M. (2002).
\newblock {\em An introduction to multiagent systems}.
\newblock John Wiley \& Sons, Chichester, England.

\end{thebibliography}
\end{document}